\documentclass[oneside,onecolumn]{article}
\usepackage{microtype}
\usepackage[english]{babel} 

\let\svthefootnote\thefootnote
\newcommand\freefootnote[1]{%
  \let\thefootnote\relax%
  \footnotetext{#1}%
  \let\thefootnote\svthefootnote%
}

\usepackage{amsthm}
\usepackage{thmtools}
\usepackage{thm-restate}

\usepackage{amsmath}

\usepackage{tabularx}

\theoremstyle{plain}
\newtheorem{theorem}{Theorem}
\newtheorem{lem}{Lemma}

\newtheorem{defn}{Definition}

\newtheorem*{rem*}{Remark}

\newtheorem{example}{Example}

\usepackage[disable,textsize=tiny]{todonotes}
\usepackage[textsize=tiny]{todonotes}

\usepackage{multirow}
\usepackage{booktabs}

\usepackage[utf8]{inputenc} 
\usepackage[T1]{fontenc}
\usepackage{url}
\usepackage{ifthen}
\usepackage{cite}
\usepackage{nicefrac}
\usepackage[hmarginratio=1:1,top=32mm,columnsep=20pt]{geometry} 

\usepackage{microtype} 

\usepackage[english]{babel} 

\usepackage{amsthm}
\usepackage{thmtools}
\usepackage{thm-restate}

\usepackage{amsmath}

\usepackage[disable,textsize=tiny]{todonotes}
\usepackage[textsize=tiny]{todonotes}

\usepackage{multirow}
\usepackage{booktabs}

\usepackage[utf8]{inputenc} 
\usepackage[T1]{fontenc}
\usepackage{url}
\usepackage{ifthen}
\usepackage{cite}
\usepackage{nicefrac}
\usepackage[hmarginratio=1:1,top=32mm,columnsep=20pt]{geometry} 

\usepackage{balance} 

\usepackage{enumitem}
\usepackage{tikz}
\usepackage{amsmath,amsfonts}
\usepackage{algorithmic}
\usepackage{algorithm}
\usepackage{array}
\usepackage[caption=false,font=normalsize,labelfont=sf,textfont=sf]{subfig}
\usepackage{textcomp}
\usepackage{stfloats}
\usepackage{url}
\usepackage{verbatim}
\usepackage{graphicx}
\usepackage{amsmath,amsfonts,bm}

\def\eqref#1{equation~\ref{#1}}

\def\1{\bm{1}}

\DeclareMathAlphabet{\mathsfit}{\encodingdefault}{\sfdefault}{m}{sl}
\SetMathAlphabet{\mathsfit}{bold}{\encodingdefault}{\sfdefault}{bx}{n}

\DeclareMathOperator*{\argmax}{arg\,max}
\DeclareMathOperator*{\argmin}{arg\,min}

\usepackage{hyperref}
\usepackage{url}
\usepackage{wrapfig}

\usepackage[utf8]{inputenc} 
\usepackage{nicefrac}       
\usepackage{textcomp}

\usepackage{microtype}
\usepackage{tcolorbox}
\usepackage{graphicx}
\usepackage{booktabs} 
\usepackage[normalem]{ulem}
\usepackage{url,multirow,comment}
\let\proof\relax

\usepackage{algorithm}
\usepackage{algorithmic}

\usepackage{mathtools}
\usepackage{enumitem}
\usepackage[capitalize,noabbrev]{cleveref}
\usepackage{booktabs}
\usepackage{caption}
\usepackage{fancyhdr}
\usepackage{graphicx}

\usepackage{bbm}
\usepackage{thmtools,thm-restate}
\usepackage{rotating}
\usepackage{pgfplots}

\title{Counterfactual Explanations for Model Ensembles Using Entropic Risk Measures}

\author{Erfaun Noorani, Pasan Dissanayake, Faisal Hamman, Sanghamitra Dutta\\
University of Maryland College Park
}

\newcommand{\BibTeX}{\rm B\kern-.05em{\sc i\kern-.025em b}\kern-.08em\TeX}

\begin{document}



\date{}
\maketitle

\let\thefootnote\relax\footnotetext{This work is accepted for publication in the proceeding of the 24th International Conference on Autonomous Agents and Multiagent Systems (AAMAS 2025), May 19 – 23, 2025, Detroit, Michigan, USA.} 

\begin{abstract}
Counterfactual explanations indicate the smallest change in input that can translate to a different outcome for a machine learning model. Counterfactuals have generated immense interest in high-stakes applications such as finance, education, hiring, etc. In several use-cases 
the decision-making process often relies on an ensemble of models rather than just one
.
Despite significant research on counterfactuals for one model, the problem of generating a single counterfactual explanation for an ensemble of models has received limited interest. Each individual model might lead to a different counterfactual, whereas trying to find a counterfactual accepted by all models might significantly increase cost (effort). 
We propose a novel strategy to find the counterfactual for an ensemble of models using the perspective of entropic risk measure. Entropic risk is a convex risk measure that satisfies several desirable properties. We incorporate our proposed risk measure into a novel constrained optimization to generate counterfactuals for ensembles that stay valid for several models. The main significance of our measure is that it provides a knob that allows for the generation of counterfactuals that stay valid under an adjustable fraction of the models. We also show that a limiting case of our entropic-risk-based strategy yields a counterfactual valid for all models in the ensemble (worst-case min-max approach). We study the trade-off between the cost (effort) for the counterfactual and its validity for an ensemble by varying degrees of risk aversion, as determined by our risk parameter knob. We validate our performance on real-world datasets. 
\end{abstract}

\section{Introduction}
The widespread adoption of machine learning models in critical decision-making, from education to finance ~\cite{dennis2018artificial,bogen2019all,chen2018fair,hao2019can}, has raised concerns about the explainability of these models~\cite{Molnar2019Interpretable,lipton2018mythos}. To address this issue, a recently-emerging category of explanations that has gained tremendous interest is: \emph{counterfactual explanation}~\cite{wachter2017counterfactual}. Given a specific data point and a model, a counterfactual explanation (also referred to as ``counterfactual'') is a feature vector leading to a different model outcome. Typically, counterfactuals are based on the \emph{closest} point on the other side of the decision boundary of the model, also referred to as the closest counterfactual
(also see surveys~\cite{zhang2022survey,Karimi_arXiv_2020,barocas2020hidden,Mishra_arXiv_2021}). For example, in automated lending, a counterfactual can inform a denied loan applicant about specific changes such as increasing collateral by 10K can lead to loan approval.

In several applications, multiple models with distinct architectures and training processes can be trained for a specific prediction task, potentially yielding different predictions for the same input. Ensemble models in machine learning combine the predictions of multiple such models to improve overall prediction (goes way back to random forests~\cite{breiman2001random}; also used for neural networks~\cite{hansen1990neural,krogh1994neural}). Ensembles aggregate insights from a set of models, often leading to more reliable outcomes. Evidence suggests averaging ensembles work because each model will make some errors independent of one another due to the high variance inherent in neural networks
with large number of parameters~\cite{hansen1990neural,krogh1994neural}. Additionally, ensembling might also be beneficial when different models capture distinct facets of the input data, akin to how multiple interviewers might offer varied perspectives on a candidate. Sometimes, ensembling several smaller models has also been found to be more useful than training one large model~\cite{kondratyuk2020ensembling}. By leveraging ensembles, machine learning systems can mitigate biases and errors inherent in individual models, enhancing performance. Ensembling techniques are also commonly employed to address the issue of predictive multiplicity where multiple equally-well-performing models lead to different predictions on certain data points~\cite{hsu2022rashomon,watson2023predictive,modelmult_black}. 


Providing recourse for model ensembles can be challenging since each individual model would lead to a different closest counterfactual. Finding a single closest counterfactual with a reasonable cost that remains valid across all models in the ensemble is nontrivial. This \emph{worst-case} approach is overly conservative leading to counterfactuals that are potentially quite far from the original point; sometimes, it may not even identify any counterfactual if there is no region where all the acceptance regions overlap. To make sure counterfactual explanations are useful and actionable to the users, we not only need them to be close but also require them to stay valid under a reasonable portion of the models within the ensemble. In general, it might even be impossible to guarantee the existence of a counterfactual that stays valid for all possible models in the ensemble. However, one might be able to ensure acceptance for a subset of models. This generates a need for an adjustable knob to obtain counterfactuals that accommodate varying fraction of models within the ensemble.

Balancing the cost and validity of counterfactuals across an ensemble of models is crucial due to potential disparities in the cost of counterfactuals across the ensemble. Understanding this trade-off will allow practitioners to tailor explanations for specific constraints and needs effectively. By providing a flexible mechanism to adjust this trade-off, machine learning systems can better manage the complexity of ensemble scenarios, ensuring that counterfactuals are both feasible and aligned with practical considerations.



\noindent \textbf{Our Contributions:} In this work, we propose a novel entropic risk measure to quantify the reliability of the counterfactual for an ensemble of models. 
Entropic risk is a convex risk measure and satisfies several desirable properties. 
Furthermore, we incorporate our proposed risk measure in the generation of reliable counterfactuals. A significance of our measure is its ability to establish a unifying connection between a worst-case (min-max optimization) approach and risk-constrained counterfactuals. 
Our proposed measure is rooted in large deviation theory and mathematical finance~\cite{follmer2002convex}. 
Our contributions can be concisely listed as follows:

\noindent\textbf{An Entropic Risk Measure for Counterfactuals in Model Ensembles:} We propose a novel entropic risk measure to quantify the reliability of counterfactuals in an ensemble. 
Our measure is convex and satisfies several desirable properties.   
It has a ``knob''-- the risk parameter-- that can be adjusted to trade off between risk-constrained and worst-case approaches. 
While risk-constrained accounts for general ensemble in an expected sense, the worst-case scenario prioritizes the worst model within the ensemble, thus having a higher cost.  

\noindent \textbf{Formulation of Constrained-Optimization to Find Counterfactuals for Model Ensembles:} Our proposed entropic risk measure enables us to obtain risk-constrained reliable counterfactuals (see constrained optimization \ref{opt-constrained_risk}). The significance of our strategy is that it enables one to tune ``how much'' a user wants to prioritize the worst model by trading off cost (effort). Since calculating expectations over ensembles, especially with infinitely many models, can be impractical, we use empirical averages instead. This approach forms the basis of our main formulation in \ref{opt-constrained_risk_p4}, which is crucial for developing our algorithm.

\noindent \textbf{Connection to Min-Max Optimization:} We show that the worst-case approaches are, in fact, a limiting case of our entropic-risk-based approach (see Theorem~\ref{thm:1}). The extreme value of the knob (risk parameter) maps our measure back to a min-max (adversarial) approach. By establishing this connection, we show that our proposed measure is not postulated and stems from the mathematical connection with worst-case analysis.

\noindent \textbf{Experimental Results:} We include an algorithm that leverages our relaxed risk measure and finds counterfactuals for model ensembles. We provide a trade-off analysis between the cost (distance) and the validity of the counterfactual on real-world datasets, namely, HELOC \cite{fico2018a}, German Credit \cite{german_credit_data}, and Adult Income \cite{adult_income_data}.



%


Notably, in agent-based systems, agents often operate in environments with incomplete or uncertain information and must make decisions that are robust to varying strategies of other agents. For example, agents make decisions based on models predicting the behavior of other agents, but these models can differ due to varying assumptions or strategies. In a similar vein, our approach uses counterfactual reasoning with an ensemble of models to explore alternative scenarios, helping agents understand potential outcomes under different conditions. By ensuring counterfactuals are robust across a range of models, agents can make more reliable decisions despite uncertainty and diverse behaviors in the system.

\subsection{Related Works} 
Counterfactuals have been extensively studied in the literature, with numerous papers exploring methodologies and applications within the context of single models (see surveys \cite{zhang2022survey,Karimi_arXiv_2020,barocas2020hidden,Mishra_arXiv_2021}). However, ensemble models are widely recognized and extensively used in machine learning for their effectiveness in improving predictive performance. By combining multiple base models—often diverse in architecture or training data—ensemble methods harness the collective wisdom of individual models to produce more accurate and reliable predictions. Popular techniques like bagging, boosting, and stacking leverage this diversity to mitigate biases, reduce variance, and enhance overall model generalization. For a survey on ensembles, we refer the readers to \cite{mienye22}.

Despite significant research on counterfactuals for one model, the problem of finding counterfactuals for an ensemble has received limited interest. A well-studied research direction involves robust counterfactuals that account for changes in models~\cite{hancox2020robustness,upadhyay2021towards,black2021consistent,hamman_2023,hamman2024robust,rawal2020can,dutta2022robust,jiang2022formalising,alvarez2018robustness, pmlr-v222-jiang24a, stkepka2024counterfactual}. Other works examine counterfactual robustness to small feature variations (noisy implementation)~\cite{bui2025coverage,dhurandhar2024trust, pmlr-v162-dominguez-olmedo22a, pawelczyk2022probabilistically,laugel2019issues, maragno2023finding, nguyen2022robust}
and and distribution shifts~\cite{rawal2020can, meyer2024verified, delaney2021uncertainty}. We refer to recent surveys on robust counterfactual explanations ~\cite{jiang2024robust,mishra2021survey}. Applying such techniques to ensembles, where the constituent models are known a priori, may be considered excessive. Unlike scenarios where the model identities are unknown or variable, our problem involves a fixed ensemble, allowing for more targeted approaches. 

Closely related is robustness under model multiplicity~\cite{pawelczyk2020counterfactual, hamman_2023,hamman2024robust, kinjo2025robust, leofante2023counterfactual}. \cite{pawelczyk2020counterfactual} suggests that counterfactuals within the data manifold are more resilient against model multiplicity compared to closest counterfactuals. \cite{hamman_2023,hamman2024robust} introduce a stability measure to quantify the robustness of counterfactuals under model multiplicity and provide probabilistic guarantees. \cite{kinjo2025robust} proposes using Pareto improvement, a multi-objective optimization to generate robust counterfactuals under model multiplicity. \cite{leofante2023counterfactual} and \cite{jiang2023recourse} propose approaches to compute counterfactuals that hold across all models within an ensemble of neural networks (or, tree-based models~\cite{parmentier2021optimal,dutta2022robust}). Our contribution lies in first developing a rigorous quantification of reliability that is specifically tailored to generate counterfactual explanations for ensembles. We use entropic risk measures that arrive with a knob to tradeoff cost and validity of counterfactuals across the models in the ensemble. Our contribution lies in developing a methodology specifically tailored to generate reliable counterfactual explanations for ensembles using entropic risk measures with a knob that trades off the cost and overall validity of counterfactuals, aiming to provide counterfactual with reasonable costs that stay valid under as many models as possible in the ensemble.

Entropic risk measure has been the cornerstone of risk-sensitive control (see \cite{jacobson1973optimal,speyer1974optimization,kumar1981optimal,james1994risk,baras1994robust,james1995robust,james1996partially,baras1998robust,noorani2021risk1,noorani2021risk2}) and risk-sensitive Markov decision processes (see \cite{howard1972risk}). The connection between risk-sensitive control and robust control has been shown in its full generality in \cite{james1994risk}, establishing that the entropic risk measure emerges from the mathematical analysis of H-infinity output robust control for general non-linear systems and has been used to trade off robustness and performance in feedback control design. Further analytical development of such mathematical analysis for financial applications has been studied extensively; see ~\cite{follmer2002convex} and references therein.

\section{Preliminaries}
Here, we provide some contextual details, definitions, and background materials, and set our notation. 
We consider machine learning models $m \in \mathcal{M}$, where $\mathcal{M}$ 
is a non-empty set of ensemble models for binary classification that takes an input value $ x \in \mathcal{X} \subseteq \mathbb{R}^d$ and outputs a probability between $0$ and $1$. Let $\mathcal{S}=\{x_i \in \mathcal{X} \}_{i=1}^n$ be a dataset of $n$ independent and identically distributed data points generated from an unknown density over $\mathcal{X}$.



\begin{defn}[Closest Counterfactual $\mathcal{C}_{p}(x,m)$]\label{def:ClosCount}
A closest counterfactual with respect to the model $m(\cdot)$ of a given point $x\in \mathbb{R}^d$ such that $m(x)<0.5$ is a point $x'\in \mathbb{R}^d$ such that $m(x')\geq0.5$ and the cost in terms of $p$-norm $\|x-x'\|_p$ is minimized. 
\begin{equation}
\mathcal{C}_{p}(x,m)=\argmin_{x'\in \mathbb{R}^d} c(x,x')
\quad \textrm{s.t.} \quad m(x')\geq0.5. \nonumber
\end{equation}  
\end{defn}

For example, norm $p=1$ results in counterfactuals with as few feature changes as possible, enforcing a sparsity constraint (also referred to as ``sparse'' counterfactuals~\cite{pawelczyk2020counterfactual}).

In this work, our \textbf{goal} is to generate a single counterfactual that is accepted by as many models as possible in the ensemble while minimizing the cost. Towards this goal, we propose an entropic risk measure as a systematic measure of the reliability of counterfactuals. Our objective involves: (i) arriving at a measure 
for the reliability of a counterfactual $x$ under a given ensemble $\mathcal{M}$, that satisfies desirable properties; (ii) establishing the connection between our entropic-risk-based approach and the worst-case approaches, and (iii) showing the algorithmic impacts of our measure by developing a constrained-optimization-based algorithm for generating counterfactuals for model ensembles based on our reliability measure which allows for a tunable knob that allows one to trade off between the cost (effort) and potential validity on multiple models.

\section{Main Results: Reliability via Entropic Risk Measure}
For a single model, the counterfactual \( x' \) would simply be the closest point to the original instance \( x \) that lies on the accepted side. We would minimize the \( \ell_2 \)-norm (i.e., the ``cost'' \( c(x, x') = \|x - x'\|_2 \) remains low). However, when we introduce the ensemble, ensuring that the counterfactual remains valid across a specified fraction of models in the ensemble can often make it move further from \( x \) to satisfy this additional requirement. This results in a higher \( \ell_2 \)-norm, increasing the cost \( c(x, x') \). When generating a counterfactual for a reference model \( m_r \) and an ensemble of models, we also seek to ensure its validity across multiple models within the ensemble, defining its *reliability*. The higher the required fraction of models that must validate the counterfactual, the more robust (reliable) it must be. However, this robustness comes at a cost: as more models must agree, the counterfactual tends to shift further from the original instance. A trade-off arises between the counterfactual’s distance from \( x \) and the robustness constraint. To balance this tradeoff, we formulate a general multi-objective optimization that hedges against the worst-case models while managing both cost and robustness, i.e.,
\begin{align*} \tag{P}\label{opt-bi}
     \min_{x'\in \mathbb{R}^d} (c(x,x'), \max_{m \in \mathcal{M}} \ell(m(x'))) \quad \textrm{s.t.} \quad m_r(x') \geq 0.5.
\end{align*}
Here $c: \mathcal X \times \mathcal X \to \mathbb R_{+}$ is the cost of changing an instance $x$ to $x'$, e.g., $c(x,x')= \|x-x'\|_p$, where $1\leq p \leq \infty$, and $\ell: \mathcal M \times \mathcal X \to R_{+}$ is a differentiable loss function that ensures that $m(x')$ is close to the desired value of $1$, e.g., $\ell(m(x))= 1-m(x)$. We denote the ensemble as \( \mathcal{M} \), where any model within the ensemble is represented by \( m \). The reference model of interest, denoted as \( m_r \), is a specific, fixed model within \( \mathcal{M} \). In \ref{opt-bi}, neither \( m \) nor \( m_r \) are treated as random variables. The second objective function $\max_{m \in \mathcal{M}} \ell(m(x'))$ is the worst-case loss over the ensemble set $\mathcal{M}$. 

To address a multi-objective optimization problem of this nature, we can seek the Pareto optimal front using established techniques, such as linear scalarization or the epsilon-constraint methods \cite{miettinen1999nonlinear}. The linear scalarization approach, for instance, entails solving
\begin{align*} \tag{P1}\label{opt-minmax}
    \min_{x'\in \mathbb{R}^d} \max_{m \in \mathcal{M}} c(x,x') + \lambda \ell(m(x')) \quad \textrm{s.t.} \quad m_r(x') \geq 0.5
\end{align*}
for different values of $\lambda$ to generate Pareto optimal solutions (e.g., a relaxed variant of this approach is employed in \cite{upadhyay2021towards}), meanwhile, the epsilon-constraint method addresses the problem by solving
\begin{align*} \tag{P2}\label{opt-constrained}
        \min_{x'\in \mathbb{R}^d} c(x,x') \quad \textrm{s.t.} \quad  \max_{m\in \mathcal{M}} \ell(m(x')) < \tau, \quad m_r(x') \geq 0.5
\end{align*}
for different values of $\tau$ (e.g., a relaxed variant of this approach is employed in \cite{hamman_2023}). 

By varying $\lambda$ in \ref{opt-minmax} or $\tau$ in \ref{opt-constrained}, different points on the Pareto front can be obtained (also see the book~\cite{miettinen1999nonlinear}). To see the equivalence of the threshold $\tau$ and the multiplier $\lambda$, note that the sensitivities of the cost $c(x,x')$ with respect to changes in the threshold $\tau$ (evaluated at the optimal $x'^{*}$) is the negative of the optimal multiplier (dual variable) $\lambda$ (for a background on multi-objective optimization, please refer to Appendix~\ref{appndx:multi-objectiveOpt} \cite{castillo2008sensitivity}), i.e, 
$\nicefrac{\partial c(x,x'^*)}{\partial \tau} = - \lambda^*.$
Each $\lambda$ and $\tau$ results in a point on the Pareto optimal front of the multi-objective optimization problem \cite{miettinen1999nonlinear,castillo2008sensitivity}. Both \ref{opt-minmax} and \ref{opt-constrained} lead to the same Pareto front, and $\lambda$ and $\tau$ can be chosen such that \ref{opt-minmax} and \ref{opt-constrained} have the same solutions. The Pareto front characterizes the trade-off between the cost and validity of the counterfactuals.

The worst-case loss $\max_{m\in \mathcal{M}} \ell(m(x'))$ hedges against the worst possible model within the ensemble, but can often lead to somewhat conservative counterfactuals, i.e., ones which are quite well within the boundary and have a high cost (distance). To mitigate this issue, we use a risk measure that allows us to \emph{hedge against the models based on their probability of occurrence}. We assume $M$ is a random model drawn from a probability distribution $P$ over the set of models $\mathcal{M}$. We propose the entropic risk measure as a quantification of reliability for counterfactuals which is defined as follows: 
\begin{defn} \label{defn:risk1}
 The entropic risk measure of model $M$ with the risk aversion parameter $\theta>0$ is denoted by $\rho^{ent}_{\theta}(\cdot)$ and is given by:
\begin{align} \label{entropic}
    \rho^{ent}_{\theta}(\ell(M(x'))):= \frac{1}{\theta} \log (\mathbb E_{M \sim P}[e^{\theta \ell(M(x'))}]), \quad \theta>0.
\end{align} 
\end{defn}
The parameter $\theta$ is called the risk parameter. A positive risk parameter results in risk-averse behavior. Hence, we refer to a positive risk parameter as the risk-aversion parameter. We show in Theorem~\ref{thm:1} that as we increase the risk-aversion parameter, our probabilistic method converges to a worst-case formulation. 
Definition~\ref{defn:risk1} allows us to reformulate our problem as follows: 
\begin{align*} \tag{P3}\label{opt-constrained_risk}
    \min_{x' \in \mathbb{R}^d} c(x,x') \quad \textrm{s.t.} \quad  \rho^{ent}_{\theta}(\ell(M(x'))) < \tau, \quad m_r(x') \geq 0.5.
\end{align*}

\subsection{Properties of Entropic Risk Measure}
Entropic risk measure is rooted in large deviation theory and is not postulated. This measure enables establishing a connection to worst-case approaches for finding counterfactuals. Taylor's expansion of the exponential shows that the entropic risk measure is the infinite sum of the moments of the distribution. Furthermore, 
it is well-known \cite{follmer2002convex} that entropic risk measure is a \emph{convex} risk measure and as such, for a positive risk parameter $\theta $$>$$0$, satisfies the properties of (1) monotonicity, (2) translation-invariance, and (3) convexity. %
\begin{enumerate} [leftmargin=*]
    \item \textbf{Monotonicity.} For $\ell(M_1(\cdot)) \geq \ell(M_2(\cdot))$,
    $$
    \rho^{ent}_{\theta}(\ell(M_1(\cdot))) \geq \rho^{ent}_{\theta}(\ell(M_2(\cdot))). 
    $$
    \item \textbf{Translation invariance.} For constant $\alpha $$\in$$ \mathbb{R}$, 
$$
         \rho^{ent}_{\theta}(\ell(M(\cdot))+\alpha) = \rho^{ent}_{\theta}(\ell(M(\cdot))) +\alpha.
$$ 
    
    \item \textbf{Convexity.} For $\alpha \in  [0,1]$,
    \begin{align*}
        \rho^{ent}_{\theta}(\alpha \ell(M_1(\cdot)) + (1-\alpha) \ell(M_2(\cdot))) \leq \alpha \rho^{ent}_{\theta}(\ell(M_1(\cdot))) + (1-\alpha) \rho^{ent}_{\theta}(\ell(M_2(\cdot))).
    \end{align*}
\end{enumerate}
For the sake of simplicity, consider the choice of cost function $\ell(M(x)) = 1-M(x)$. Then, the monotonicity implies that a model with greater output probabilities has less risk. The translation invariance implies that adding a constant to the output of the predictor effectively reduces the risk by the same amount. The convexity is quite desirable since it means that the risk for a combined model is lower than the risk for the two of them individually.

To gain a deeper understanding of the risk constraint described in \ref{opt-constrained_risk}, we examine distributions characterized by their analytical Moment Generating Functions (MGFs). Two notable examples are the Uniform and truncated Gaussian distributions. 
%
For simplicity, we use the cost function $\ell(M(x')) $$=$$ 1 $$-$$ M(x')$. 
In our formulation, this loss function is minimized, encouraging a counterfactual with a higher predicted value. When using this specific cost function, any value of the threshold $\tau$ outside the interval $[0, 1]$ renders the problem infeasible. Given these choices for the cost and model distribution, we provide the explicit form of the constraint in \ref{opt-constrained_risk}.


\begin{example} \label{example:uniform}
    Let the distribution of the output of the models in the ensemble at the counterfactual point, $M(x')$, follow a uniform distribution on a $\delta$-ball around the output of a specific model $m(x')$, i.e., $M(x') \sim \mathcal U [m(x')- \delta, m(x')+ \delta]$ for some $\delta>0$. With these choices, the constraint in \ref{opt-constrained_risk} becomes:
    \begin{align*}
        m(x') > (1-\tau) + K_{\delta,\theta}, \quad         K_{\delta,\theta} := \frac{1}{\theta} \log(\frac{e^{\theta \delta}-e^{-\theta \delta}}{2\theta \delta}).
    \end{align*}
\end{example}

For the Uniform distribution, due to the monotonicity of $K_{\delta,\theta}$ with respect to $\theta$, as the value of $\theta$ increases, a higher value of $m(x')$ is required to satisfy the constraint. It can be verified that $K_{\delta,\theta}$ in limit of $\theta \to \infty$ is $\delta$. Given this, for the case when $\theta \to \infty$, our constraint becomes $m(x') > 1-\tau+\delta$. As the value of $\theta$ approaches to $0$, $K_{\delta,\theta}$ approaches $0$ and the constraint becomes $m(x')>(1-\tau)$, i.e., finding counterfactual $x'$ with just high $m(x')$.

\begin{example} [Truncated Gaussian] \label{example:truncated_gaussian}
Let the distribution of the output of the models in the ensemble at the counterfactual point, $M(x')$, follow a truncated Gaussian distribution with a mean equal to the output of the original model $m(x')$ and a variance of $\sigma^2$ that lies between $0$ and $1$. With these choices, the constraint in \ref{opt-constrained_risk} becomes:
    \begin{align*}
        m(x') > (1-\tau )+  \theta \frac{\sigma^2}{2} + \frac{1}{\theta} \log (K_{\theta}),\quad K_{\theta}:= \frac{\Phi(\beta + \sigma\theta)-\Phi(\alpha + \sigma\theta)}{\Phi(\beta)-\Phi(\alpha)}
    \end{align*}
where $\alpha := \frac{-\mu}{\sigma}$ and $\beta := \frac{1-\mu}{\sigma}$ and $\Phi(x) = \nicefrac{1}{2} (1+\operatorname {erf}(x/\sqrt 2))$. The error function, denoted by $\operatorname {erf}$, is defined as 
$
    {\displaystyle \operatorname {erf} z={\nicefrac {2}{\sqrt {\pi }}}\int _{0}^{z}e^{-t^{2}}\,\mathrm {d} t.}
$
\end{example}
As the $\theta$ approaches $0$, our constraint becomes $m(x')>1-\tau$. As the value of $\theta$ increases, greater weight is placed on the variance term, emphasizing its importance. 
In both examples, when the distributions are unknown, determining the precise threshold for model output to satisfy the constraint becomes challenging. This is because higher values are more conservative (less risky), but incur higher costs. To address this challenge, we must devise techniques that do not rely on the explicit knowledge of the distribution, as explored further in the next subsections.

\subsection{Connection of Entropic-Risk-Based Approach with Worst-Case Approach}
We first establish the connection between our risk-based and the worst-case formulation (getting accepted by all models in the ensemble). 
The following theorem shows that the worst-case approach is the limiting case of our risk-based method as $\theta \to \infty$
. 

\begin{restatable}{theorem}{thmopt} 
\label{thm:1}
    In the limit as the risk-aversion parameter $\theta$ approaches infinity, the optimization \ref{opt-constrained_risk}, which involves constraining the entropic risk measure associated with the reliability of models within an ensemble, asymptotically converges to the optimization problem  \ref{opt-constrained}, where the constraint pertains to the robustness of the worst model within the same ensemble.
\end{restatable}
Theorem~\ref{thm:1} shows how the entropic risk measure provides a single parameter (knob) that determines the risk-aversion of the counterfactual and can be used to study the effect of risk-aversion on the behavior of algorithms that generate reliable counterfactuals for an ensemble.\\
\noindent \textbf{Proof Sketch of Theorem\ref{thm:1}
:} We discuss the proof in Appendix~\ref{appndx:proofs}. The proof uses 
Vardhan's Lemma presented here. Such connections have been shown in the context of robust and risk-sensitive control and, more recently, risk-sensitive reinforcement learning. 

\begin{restatable}{lem}{lemmadual} \label{lemma:dual} \cite{follmer2002convex} Let $X$ be a random variable. The entropic risk measure is a convex risk measure and as such has a dual representation with the risk aversion parameter $\theta>0$ is given by
    \begin{align*}
    \rho_{\theta}^{{{\mathrm  {ent}}}}(X) &={\frac  {1}{\theta }}\log \left({\mathbb  {E}_{X \sim P}}[e^{{\theta X}}]\right) = \sup _{Q \ll P}\left\{E_{Q}[X]-{\frac  {1}{\theta }}D(Q|P)\right\}\
\end{align*}
where $D(Q|P):=\mathbb E_{Q}\left[\log{\nicefrac{dQ}{dP}}\right]$ is the Kullback-Libeler (KL) divergence between distributions $P$ and $Q$, and $Q \ll P$ denotes the distribution Q is absolutely continuous with respect to $P$. 
\end{restatable}

\subsection{Formulation of the Constrained Optimization}
We substitute the expectation in the risk measure with a computable empirical mean. This allows us to reformulate the problem as
\begin{align*} \tag{P4}\label{opt-constrained_risk_p4}
    &\min_{x' \in \mathbb{R}^d} c(x,x') \\ &\textrm{s.t.} \quad {\frac  {1}{\theta}}\log \left(  \frac{1}{N} \sum_{i=1}^{N} e^{{(1-\theta m_i(x'))}} \right) < \tau, \quad m_r(x') \geq 0.5,
\end{align*}
where $m_i$'s are the sample models from the ensemble $\mathcal{M}$.

\section{Experimental Results} \label{sec:simulations}
In this section, we experimentally demonstrate the effect of entropic risk minimization on generating counterfactual explanations. To this end, we observe that the risk aversion parameter $\theta$ plays a key role in the cost-validity trade-off. 

\noindent \textbf{Experimental Setup:} We consider an ensemble of 20 models, each with three 128-neuron hidden layers and ReLU activations. The models are trained employing the Adam optimizer for 200 epochs with a batch size of 32. The same model architecture and hyperparameters were used for all the datasets, since it yielded satisfactory levels of accuracy on all of them. We evaluate the proposed method over three publicly available datasets namely HELOC~\cite{fico2018a}, German Credit~\cite{german_credit_data} and Adult Income~\cite{adult_income_data} (see Appendix \ref{appndx:EX} for details). Each dataset is split into a training set and a test set. To generate our ensemble $\mathcal{M}$, we train each model $m_i$ on a slightly different subset of the training split, generated by dropping $k$ (a hyperparameter) randomly selected data points prior to training the model. The test split is used for evaluating model accuracies as well as for generating the counterfactuals. Table \ref{tab:exp_setup} summarizes dataset-specific details.
\begin{table}[!htbp]
\caption{Experimental setup: Standard deviations are given in parenthesis when applicable.}
\label{tab:exp_setup}
\centering
\begin{small}
\begin{tabular}{cccc} 
\toprule
Property & HELOC & GERMAN & ADULT\\
\midrule
Training set size & 7844 & 670 & 15081 \\
Test set size & 2615 & 330 & 15081 \\
\# Rejected instances & 889 & 186 & 10216 \\
\# Dropped Points $(k)$ & 1000 & 100 & 1000 \\
Average model accuracy & 0.66 (0.01) & 0.72 (0.02) & 0.82 (0.002)
\\
\bottomrule
\end{tabular}
\end{small}
\end{table} 

 \textbf{Algorithm:} In the experiments, we solve \ref{opt-constrained_risk_p4} through a two step process based on gradient descent. First, an ordinary counterfactual $x'$ is generated for a randomly selected reference model $m_{r}$ from the ensemble, using an existing counterfactual generating method ($\ell_1-$norm closest counterfactual in our case). Then the counterfactual is updated until the entropic risk constraint $\rho_\theta^{\text{ent}}(x') < \tau$ is satisfied. This is done through a gradient descent process $x' \leftarrow x' - \eta \nabla_{x'}\rho_\theta^{\text{ent}}(x')$. Counterfactuals are generated only for the instances that were rejected under the said randomly selected model. Note that for some instances, the counterfactual generation method fails to render counterfactuals due to the finite number of gradient descent iterations. A workaround for this issue is to experiment with different hyperparameters such as the gradient descent step size $\eta$ and the maximum number of iterations. Algorithm \ref{alg:cf_generation} presents the counterfactual generation steps concisely.
\begin{algorithm}
\caption{Entropic risk based counterfactual generation}
\label{alg:cf_generation}
\begin{algorithmic}
    \REQUIRE Input instance $x$, Model ensemble $\mathcal{M}$, $\theta > 0, \tau>0$, Gradient descent step size $\eta$, max\textunderscore iter $\in \mathbb{Z}^+$.
    \STATE Randomly select $\Tilde{m}\in \mathcal{M}$.
    \STATE Generate ordinary counterfactual $x' \gets \mathcal{C}_p(x, \Tilde{m})$.
    \STATE Initialize $i \gets 0$.
    \WHILE{$\rho_\theta^{\text{ent}}(x') \geq \tau$ \AND $i<$ max\textunderscore iter}
        \STATE $x' \gets x' - \eta \nabla_{x'} \rho_\theta^{\text{ent}}(x')$
        \STATE $i \gets i + 1$
    \ENDWHILE
    \IF {$\rho_\theta^{\text{ent}}(x') < \tau$}
        \RETURN $x'$ and exit
    \ELSE
        \RETURN Error (Invalid counterfactual) and exit
    \ENDIF
\end{algorithmic}
\end{algorithm}

\textbf{Metrics:} We are interested in observing the trade-off between a counterfactual being easy to achieve (low cost) and being valid under an ensemble decision, e.g., majority vote (high validity). In this regard, for each set of parameters $\theta$ and $\tau$, we compute the two metrics: (i) \textit{Cost}: the $\ell_1-$distance between the counterfactual and the corresponding original instance (ii) \textit{Validity}: the ratio of models in the ensemble with respect to which the counterfactual is valid. These metrics are averaged over all the counterfactuals that satisfy the risk constraint. For comparison, we also include the results for the case $\tau=1$ which is equivalent to not having any risk constraints (i.e., non-robust). In addition, we compute the average wall clock time taken to generate a counterfactual under each set of parameters $\theta$ and $\tau$, which is an indicator of the difficulty in generating an explanation that satisfies the risk constraint.

\textbf{Hyperparameter Selection:} Values for the gradient descent step size $\eta$ and the number of maximum iterations were selected empirically such that the algorithm would converge to a solution in a reasonable time for most of the input instances. Ranges for $\theta$ and $\tau$ were selected in a dataset-specific manner. For instance, for HELOC dataset, $\theta\in\{0.1, 1.0, 10.0\}$ results $\rho_\theta^{\text{ent}}(x') < 0.8$ for most of the input instances. Achieving $\rho_\theta^{\text{ent}}(x') < 0.1$ was seen to be difficult. These values were slightly different for the other datasets.

\begin{figure}[t]
    \centering
\includegraphics[width=0.8\columnwidth]{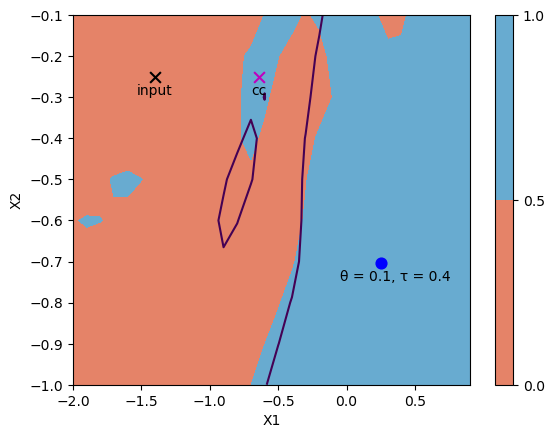}  \caption{A 2D visualization of the proposed method. Black cross denotes a rejected input instance. Magenta cross is the closest counterfactual (``cc'') in terms of $\ell_1-$distance, generated w.r.t. reference model $m_r$. Blue dot represents the entropic-risk-based counterfactual generated with $\theta=0.1$ and $\tau=0.4$. Purple line is the decision boundary of another model $m\neq m_r$ in the ensemble. Observe that even though the closest counterfactual is valid under $m_r$, it is rejected under the other model $m$. In contrast, the entropic-risk-based counterfactual remains valid w.r.t. both models. 
\label{fig:synthetic}} %
\end{figure}

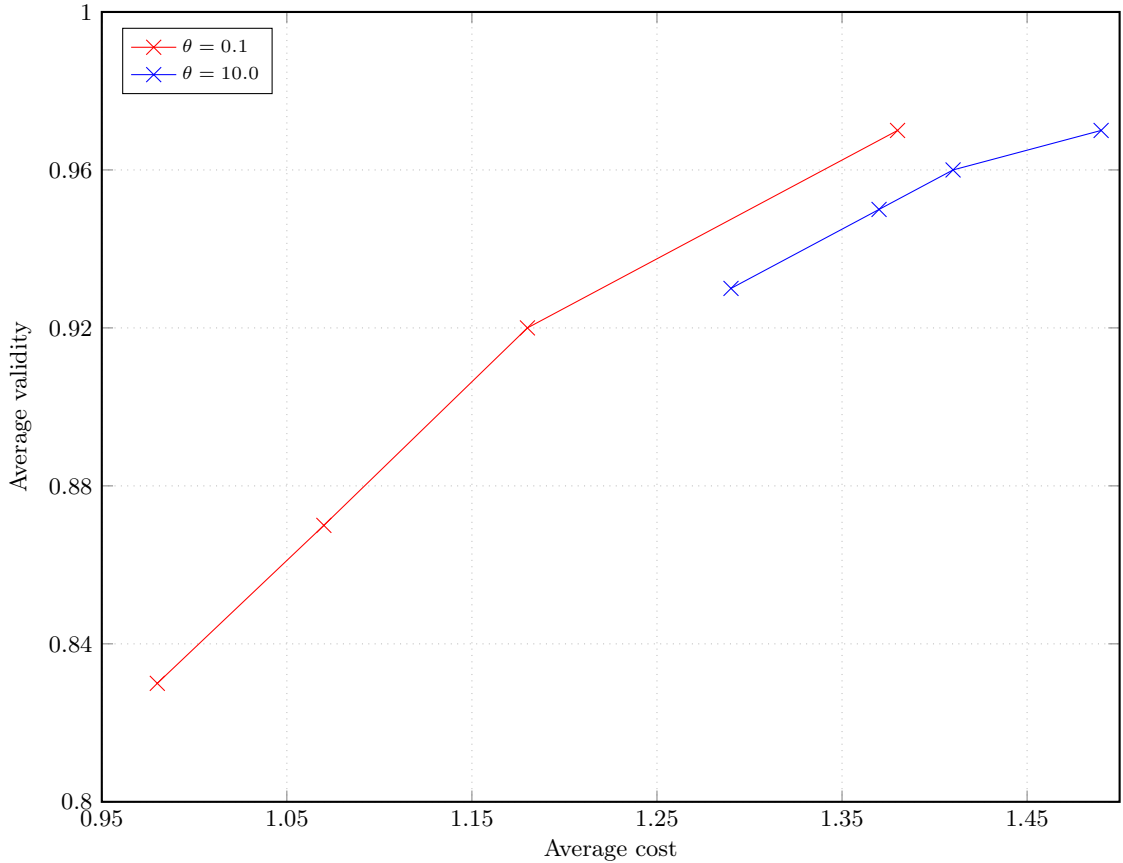
\begin{figure}[t]
\centering
\begin{tikzpicture}
\begin{axis}[
    xlabel={Average cost},
    ylabel={Average validity},
    width=\linewidth,
    height=0.8\linewidth,
    xmin=0.95, xmax=1.5,
    ymin=0.8, ymax=1.,
    xtick={0.95, 1.05, 1.15, 1.25, 1.35, 1.45},
    ytick={0.8, 0.84, 0.88, 0.92, 0.96, 1.0},
    xticklabel style={font=\small},
    yticklabel style={font=\small},
    xlabel style={font=\small, yshift=1ex, align=center},
    ylabel style={font=\small, yshift=-1ex, align=center, rotate=0},
    legend style={at={(0.02,0.98)}, anchor=north west, font=\scriptsize, fill=white},
    legend cell align={left},
    axis line style={thick},
    grid style={dotted},
    major grid style={gray!50},
    minor grid style={gray!20},
    grid=both,
]
\addplot[mark=x, mark size=4pt, red] coordinates {
    (1.38, 0.97)
    (1.18, 0.92)
    (1.07, 0.87)
    (0.98, 0.83)
};
\addlegendentry{$\theta=0.1$}
\addplot[mark=x, mark size=4pt, blue] coordinates {
    (1.49, 0.97)
    (1.41, 0.96)
    (1.37, 0.95)
    (1.29, 0.93)
};
\addlegendentry{$\theta=10.0$}

\end{axis}
\end{tikzpicture}
\caption{Cost-validity trade-off curves for different $\theta$ values on HELOC dataset. Each point on a given curve corresponds to a distinct $\tau\in\{0.1, 0.3, 0.5, 0.7\}$. Cost and validity increase monotonically with increasing $\tau$.
}
\label{fig:trade-off}
\end{figure}

\textbf{Results and Discussion:} 
In addition to the real-world data, we conducted a synthetic experiment with a 2D dataset, to facilitate easy visualization. Figure \ref{fig:synthetic} demonstrates the results corresponding to one of the input instances. Tables \ref{tab:heloc_averages}, \ref{tab:gc_averages}, and \ref{tab:ai_averages} present the results of the experiment for HELOC, German Credit, and Adult Income datasets, respectively. Observe that the validity increases with increasing $\theta$ for a given value of $\tau$, indicating how $\theta$ facilitates a smooth trade-off between the two metrics. Figure \ref{fig:trade-off} visualizes this trade-off. Tables 
\ref{tab:heloc_compute_time}, \ref{tab:gc_compute_time}, and \ref{tab:ai_compute_time} show the average wall clock time taken to generate an explanation, which can be considered as a proxy for the difficulty of generating a valid counterfactual. Notice how the time decreases with increasing $\tau$ for each value of $\theta$. Furthermore, notice that when the risk constraint is inactive (i.e., when $\tau=1$, the averaged cost and validity values are almost the same for all values of $\theta$.
\setlength{\tabcolsep}{2pt}
\begin{table*}[!htbp]
\caption{Experimental results for HELOC dataset. Standard deviations are given in parenthesis.}
\label{tab:heloc_averages}
\centering
\begin{scriptsize}
\begin{tabular}{ccccccccccc} 
\toprule
 \multirow{2}{*}{$\theta$} & \multicolumn{2}{c}{$\tau=0.1$} & \multicolumn{2}{c}{$\tau=0.3$} & \multicolumn{2}{c}{$\tau=0.5$} & \multicolumn{2}{c}{$\tau=0.7$} & \multicolumn{2}{c}{$\tau=1.0$} \\
\cmidrule(lr){2-3} \cmidrule(lr){4-5} \cmidrule(lr){2-3} \cmidrule(lr){4-5} \cmidrule(lr){6-7} \cmidrule(lr){8-9} \cmidrule(lr){10-11}
 & COST & VAL. & COST & VAL. & COST & VAL. & COST & VAL. & COST & VAL. \\
\midrule
0.1 & 1.38 (1.29) & 0.97 (0.06) & 1.18 (1.24) & 0.92 (0.11) & 1.07 (1.22) & 0.87 (0.15) & 0.98 (1.22) & 0.83 (0.19) & 0.89 (1.22) & 0.76 (0.24) \\

1.0 & 1.39 (1.27) & 0.97 (0.06) & 1.22 (1.24) & 0.93 (0.10) & 1.10 (1.24) & 0.90 (0.13) & 1.00 (1.22) & 0.84 (0.18) & 0.89 (1.22) & 0.76 (0.25) \\

10.0 & 1.49 (1.22) & 0.97 (0.09) & 1.41 (1.23) & 0.96 (0.09) & 1.37 (1.24) & 0.95 (0.11) & 1.29 (1.25) & 0.93 (0.12) & 0.88 (1.23) & 0.75 (0.25) \\
\bottomrule
\end{tabular}
\end{scriptsize}
\end{table*} 
\begin{table*}
\caption{Experimental results for German Credit dataset. Standard deviations are given in parentheses.}
\label{tab:gc_averages}
\centering
\begin{scriptsize}
\begin{tabular}{ccccccccccc} 
\toprule
 \multirow{2}{*}{$\theta$} & \multicolumn{2}{c}{$\tau=0.3$} & \multicolumn{2}{c}{$\tau=0.5$} & \multicolumn{2}{c}{$\tau=0.7$} & \multicolumn{2}{c}{$\tau=0.9$} & \multicolumn{2}{c}{$\tau=1.0$} \\
\cmidrule(lr){2-3} \cmidrule(lr){4-5} \cmidrule(lr){2-3} \cmidrule(lr){4-5} \cmidrule(lr){6-7} \cmidrule(lr){8-9} \cmidrule(lr){10-11}
 & COST & VAL. & COST & VAL. & COST & VAL. & COST & VAL. & COST & VAL. \\
\midrule
0.1 & 2.20 (1.49) & 0.87 (0.07) & 1.73 (1.48) & 0.77 (0.13) & 1.45 (1.41) & 0.66 (0.20) & 1.16 (1.38) & 0.59 (0.28) & 1.11 (1.38) & 0.58 (0.29) \\

1.0 & 2.39 (1.51) & 0.90 (0.05) & 1.91 (1.47) & 0.82 (0.10) & 1.55 (1.44) & 0.71 (0.16) & 1.20 (1.38) & 0.60 (0.27) & 1.11 (1.38) & 0.58 (0.29) \\

10.0 & 3.46 (1.49) & 1.00 (0.00) & 3.39 (1.55) & 1.00 (0.00) & 3.03 (1.51) & 0.97 (0.02) & 1.71 (1.47) & 0.77 (0.12) & 1.13 (1.38) & 0.59 (0.29) \\
\bottomrule
\end{tabular}
\end{scriptsize}
\end{table*} 
\begin{table*}
\caption{Experimental results for Adult Income dataset. Standard deviations are given in parentheses.}
\label{tab:ai_averages}
\centering
\begin{scriptsize}
\begin{tabular}{ccccccccc} 
\toprule
 \multirow{2}{*}{$\theta$} & \multicolumn{2}{c}{$\tau=0.5$} & \multicolumn{2}{c}{$\tau=0.7$} & \multicolumn{2}{c}{$\tau=0.9$} & \multicolumn{2}{c}{$\tau=1.0$} \\
\cmidrule(lr){2-3} \cmidrule(lr){4-5} \cmidrule(lr){2-3} \cmidrule(lr){4-5} \cmidrule(lr){6-7} \cmidrule(lr){8-9}
 & COST & VAL. & COST & VAL. & COST & VAL. & COST & VAL. \\
\midrule

0.1 & 1.07 (3.13) & 0.95 (0.14) & 1.02 (3.13) & 0.91 (0.20) & 1.01 (3.13) & 0.89 (0.22) & 1.00 (3.14) & 0.89 (0.23) \\

1.0 & 1.08 (3.13) & 0.96 (0.12) & 1.03 (3.13) & 0.92 (0.19) & 1.01 (3.13) & 0.89 (0.22) & 1.00 (3.14) & 0.89 (0.23) \\

10.0 & 1.15 (3.12) & 0.99 (0.07) & 1.11 (3.13) & 0.96 (0.14) & 1.03 (3.14) & 0.91 (0.20) & 1.00 (3.14) & 0.88 (0.24) \\
\bottomrule
\end{tabular}
\end{scriptsize}
\end{table*}

Moreover, we observe the effect of ensemble size $N$ on the validity of a counterfactual, for a fixed set of parameters $\theta$ and $\tau$. Intuitively, given that the model diversity is constant, the smaller the ensemble size the lower the chances of getting invalidated. Hence, when all the other parameters are constant, the validity should increase with reducing ensemble size. This is indeed the case observed empirically as reported in Table \ref{tab:heloc_ensemble_size}. Note how validity increases when the number of models in the ensemble is reduced from 20 to 10, corresponding to each value of $\tau$.

\begin{table}
\caption{Average wall clock time (in milliseconds) taken to generate a counterfactual -- HELOC dataset.}
\label{tab:heloc_compute_time}
\centering
\begin{small}
\begin{tabular}{ccc} 
\toprule
$\theta$ & $\tau=0.7$ (robust) & $\tau=1.0$ (non-robust) \\
\midrule
0.1 & 646 & 524 \\
1.0 & 670 & 525 \\
10.0 & 1180 & 520 \\
\bottomrule
\end{tabular}
\end{small}
\end{table}

\begin{table}[t]
\caption{Average wall clock time (in milliseconds) taken to generate a counterfactual -- German Credit dataset.}
\label{tab:gc_compute_time}
\centering
\begin{small}
\begin{tabular}{ccc} 
\toprule
$\theta$ & $\tau=0.7$ (robust) & $\tau=1.0$ (non-robust) \\
\midrule
0.1 & 1605 & 972 \\
1.0 & 1843 & 960 \\
10.0 & 4539 & 941 \\
\bottomrule
\end{tabular}
\end{small}
\end{table}

\begin{table}[t]
\caption{Average wall clock time (in milliseconds) taken to generate a counterfactual -- Adult Income dataset.}
\label{tab:ai_compute_time}
\centering
\begin{small}
\begin{tabular}{ccc} 
\toprule
$\theta$ & $\tau=0.7$ (robust) & $\tau=1.0$ (non-robust) \\
\midrule
0.1 & 5278 & 5189 \\
1.0 & 5277 & 5181 \\
10.0 & 5379 & 5191 \\
\bottomrule
\end{tabular}
\end{small}
\end{table}

\begin{table}[t]
\caption{Effect of ensemble size $N$. Values shown for HELOC dataset with $\theta=1.0$.}
\label{tab:heloc_ensemble_size}
\centering
\begin{small}
\begin{tabular}{ccccc} 
\toprule
\multirow{2}{*}{$\tau$} & \multicolumn{2}{c}{N=10} & \multicolumn{2}{c}{N=20} \\
\cmidrule(lr){2-3} \cmidrule(lr){4-5}
& COST & VAL. & COST & VAL. \\
\midrule
0.1 & 1.38 (1.25) & 0.96 (0.11) & 1.39 (1.27) & 0.97 (0.06) \\
0.3 & 1.22 (1.24) & 0.93 (0.12) & 1.22 (1.24) & 0.93 (0.10) \\
0.5 & 1.11 (1.23) & 0.89 (0.14) & 1.10 (1.24) & 0.90 (0.13) \\
0.7 & 1.01 (1.22) & 0.83 (0.19) & 1.00 (1.22) & 0.84 (0.18) \\
\bottomrule
\end{tabular}
\end{small}
\end{table}

\section{Conclusion \& Limitations}
With our entropic risk measure, we showed that the risk-aversion parameter can be adjusted to balance the cost and validity of counterfactuals by considering the impact of the worst model. We showed that the worst-case approach is a limiting case of our approach based on entropic risk measures. This establishes the connection between our approach and a worst-case approach and explains the nature of the counterfactuals generated by our algorithm. Our research also makes a broader connection between the field of explainability and multi-objective optimization through the lens of risk measures.

Another related research direction is model reconstruction~\cite{dissanayake2024model} where it has been found that counterfactuals lead to more efficient model reconstruction since they are quite close to the decision boundary. In this context, such strategies of generating counterfactuals for ensembles could also have potential applications in defending against such extraction attacks since they are less uniquely tied to a particular model, enhancing privacy.


The integration of machine learning systems into our daily lives has wide-ranging and complex implications. These implications range from economic to societal to ethical and legal considerations, necessitating a comprehensive approach to address the sociotechnical evolution driven by machine learning.  While our current work represents a step towards trustworthy adoption, counterfactual explanations also suffer from a multitude of other limitations such as fairness, actionability, and personalization~\cite{sharma2019certifai,ley2022global,Karimi_arXiv_2020}
Consider this scenario, when examining a loan approval, a counterfactual suggesting an increase in the value of the applicant's collateral might be perceived as more preferable for an applicant as opposed to a counterfactual suggesting an increase in education level even if they might have the same $l_1$ cost. Therefore, in our future work, we will explore approaches that incorporate additional metrics beyond explainability and reliability to generate counterfactuals, addressing other relevant considerations. 

By ensuring the reliability and trustworthiness of counterfactuals from both user and institutional perspectives, we can foster greater trust in machine learning systems, leading to broader economic benefits and reliable adoption of machine learning in high-stakes applications. However, it is important to recognize that achieving the reliability of counterfactuals for ensembles requires solving computationally more expensive constrained optimization problems compared to the closest counterfactual for a single model. Therefore, future efforts should focus on devising more computationally efficient techniques to overcome this challenge and ensure the sustainability of counterfactual generation approaches.

\appendix
\section{Proof of Theorem \ref{thm:1}} \label{appndx:proofs}

\thmopt*


The proof of Theorem~\ref{thm:1} uses the results in Lemma~\ref{lemma:dual} and \ref{lemma:sup}.

\lemmadual*


Note that $Q$ is absolutely continuous with respect to $P$ if $Q(x)=0$ when $P(x)=0$. This assumption ensures that the KL divergence is finite. Then, we have,
\begin{equation}
    \lim_{\theta \to \infty} \rho ^{{{\mathrm  {ent}}}}(X)=\sup_{Q \ll P}\left\{E_{Q}[X]\right\}. \label{eq:limit}
\end{equation}
For simplicity, we let both $Q(\tilde{m})>0$ and $P(\tilde{m})>0$ over the set of models $\mathcal{M}$ which is a compact and bounded set. Next, we show the following result.




\begin{lem} \label{lemma:sup}
    Let $Q$ be any probability distribution over the set of models $\mathcal{M}$ such that $Q(\tilde{m})>0$ everywhere, and $\mathcal{M}$ be a compact and bounded set. Then we have,
    \begin{align*}
        \sup_{{Q}} \mathbbm E_{Q}[\ell (M)] = \max_{m_i \in \mathcal{M}} \ell(m_i)
    \end{align*}
\end{lem}

\proof{
We prove the equality by establishing two directions of the inequality.
First, we note that the expected value of a set of values is always less than or equal to its maximum value. Thus, 
\begin{align*}
    \mathbbm E_Q[\ell(M)] \leq \max_{m \in \mathcal{M}} \ell(m), \quad \forall Q
\end{align*}
Since it holds for all $Q$'s we have
\begin{align} 
    \sup_{Q} \mathbbm E_Q[\ell(M)] \leq \max_{m \in \mathcal{M}} \ell(m)
\end{align}
To prove the reverse direction, let $Q_{m}$ be a probability distribution such that
\begin{align*}
    Q_{m}(\tilde{m})= \begin{cases} 
      1 - \delta & \tilde{m}=m \\
      \delta_{\tilde{m}} & \tilde{m} \neq m
   \end{cases}
\end{align*}
where $\delta_{\tilde{m}} \neq 0$, for all $\tilde{m} $$\in$$ \mathcal M$ and $\delta $$=$$ \sum_{\tilde{m} \in \mathcal M, \tilde{m} \neq m} \delta_{\tilde{m}}$. Then, we have 
\begin{align*}
    \mathbbm E_{Q_{m}}[\ell(M)] = (1-\delta) \ell(m) + \sum_{\tilde{m} \in \mathcal M, \tilde{m} \neq m} \delta_{\tilde{m}} \ell(\tilde{m}), \quad \forall m
\end{align*}
Thus,
\begin{align*}
    \sup_Q \mathbbm E[\ell(M)] &\geq \mathbbm \mathbbm E_{Q_{m}}[\ell(M)] \\
    &= (1-\delta) \ell(m) + \sum_{\tilde{m} \in \mathcal M, \tilde{m} \neq m} \delta_{\tilde{m}} \ell(\tilde{m}), \quad \forall m
\end{align*}
Let $m^* = \argmax_{m} \ell(m)$. Then we have,
\begin{align*}
    \sup_Q \mathbbm E[\ell(M)] \geq (1-\delta) \ell(m^*) + \sum_{\tilde{m} \in \mathcal M, \tilde{m} \neq m^*} \delta_{\tilde{m}} \ell(\tilde{m}) 
\end{align*}

By noting that $\delta$ can be made arbitrarily small, we have
\begin{align*}
    \sup_Q \mathbbm E[\ell(M)] \geq \max_{m \in \mathcal{M}} \ell(m) - \epsilon{(\delta)}
\end{align*}
for an arbitrarily small $\epsilon{(\delta)}>0$. Thus the result holds. 

The set $\mathcal M$ needs to be such that the maximum exists, e.g., a bounded and compact set.
}

Now using Lemma~\ref{lemma:sup}, we have
\begin{align*}
    \lim_{\theta \to \infty} \rho^{ent}_{\theta}(\ell(m(x'))) &:= \frac{1}{\theta} \log (\mathbb E_{M \sim P}[e^{\theta \ell(M(x)}]) \\ &\overset{(a)}{=} \sup _{{Q\in {\mathcal  {M}}_{1}}}\left\{E_{Q}[\ell(M(x))]\right\} \\ &\overset{(b)}{=} \sup_{m \in \mathcal{M}} \ell(m(x')),
\end{align*}
where (a) holds since $\lim_{\theta \to \infty} \rho ^{{{\mathrm  {ent}}}}(X)=\max_{Q \ll P}\left\{E_{Q}[X]\right\}$ as shown in \eqref{eq:limit} and (b) follows from Lemma \ref{lemma:sup}. 
\section{Background on Multi-Objective Optimization}\label{appndx:multi-objectiveOpt}
Consider a non-linear programming problem with inequality constraints such as:
\begin{align*}
\min_{x'} c(x,x') \qquad \textrm{subject to:}
\quad R(x, x') \leq \tau
\end{align*}
where $c$ and $R$ are regular enough for the mathematical developments to be valid over
the feasible region. It is also assumed that the problem has an optimum. Then the sensitivities of the objective function with respect to the threshold $\tau$ can be calculated using the following theorem:

\begin{theorem} \cite{castillo2008sensitivity}
Assume that the solution of the above optimization problem is a regular point and that no degenerate inequality constraints exist. Then, the sensitivity of the objective function with respect to the parameter a is given by the gradient of the Lagrangian function
\begin{align*}
 L = c(x, x') + \lambda^T (R(x, x') - \tau)
\end{align*}
with respect to $\tau$ evaluated at the optimal solution $x^*$, i.e.,
\begin{align*}
   \frac{\partial c(x,x^*)}{\partial \tau} = \nabla_{\tau} L = -\lambda^* 
\end{align*}

\end{theorem}
where $\lambda^*$ is the dual optimal solution.
This shows how much the objective function value $c$ changes when parameter $\tau$ changes. 


\section{Experiments} \label{appndx:EX}
\subsection{Datasets}

\textbf{HELOC.} The FICO HELOC ~\cite{fico2018a} dataset contains anonymized information about a home equity line of credit applications made by homeowners in the US, with a binary response indicating whether or not the applicant has ever been more than 90 days delinquent for a payment. It can be used to train a machine learning model to predict whether the homeowner qualifies for a line of credit or not. The dataset consists of 10459 rows and 40 features, which we have normalized to be between zero and one.
\\

\noindent\textbf{German Credit.} The German Credit dataset~\cite{german_credit_data} comprises 1000 entries, each representing an individual who has taken a credit from a bank. These entries are characterized by 20 categorical features, which are used to classify each person as a good or bad credit risk. To prepare the dataset, we one-hot encoded the data and normalized it such that all features fall between zero and one.
\\

\noindent\textbf{Adult Income.} The Adult Income \cite{adult_income_data} dataset comprises entries for 48842 individuals with a collection of 14 features for each of them. The target is a binary variable that indicates whether the individual has an income exceeding \$50,000 or not. All the features are normalized to lie between zero and one.

\section*{ACKNOWLEDGMENTS} This work was supported in part by Northrop Grumman Seed Grant and NSF CAREER Award 2340006.

\bibliography{main_refs}
\bibliographystyle{unsrt}

\end{document}